\documentclass[runningheads]{llncs}
\usepackage[T1]{fontenc}
\usepackage{graphicx}
\usepackage[bookmarks=false,hidelinks]{hyperref}
\usepackage{color}

\begin{document}
\title{Optimizing Multimodal Language Models through Attention-based Interpretability}
\titlerunning{Optimizing MLMs through Attention-based Interpretability}
%
\author{Alexander Sergeev\inst{1}\orcidID{0009-0002-4103-842X} \and \\
Evgeny Kotelnikov\inst{1}\orcidID{0000-0001-9745-1489}}
\authorrunning{A. Sergeev, E. Kotelnikov}
%
\institute{European University at Saint Petersburg, St Petersburg, Russia \\
\email{\{sergeev.alexander0,kotelnikov.ev\}@gmail.com}}
\maketitle              
\begin{abstract}
Modern large language models become multimodal, analyzing various data formats like text and images. While f\/ine-tuning is ef\/fective for adapting these multimodal language models (MLMs) to downstream tasks, full f\/ine-tuning is computationally expensive. Parameter-Ef\/f\/icient Fine-Tuning (PEFT) methods address this by training only a small portion of model weights. However, MLMs are dif\/f\/icult to interpret, making it challenging to identify which components are most ef\/fective for training to balance ef\/f\/iciency and performance.

We propose an attention-based interpretability method for MLMs by analyzing attention scores relative to image tokens. The core idea is to identify attention heads that focus on image key objects. We utilize this information to select optimal model components for PEFT in multimodal models. Our contributions include a method for identifying attention heads associated with image key objects, its application to PEFT for image captioning, and the creation of a new dataset containing images, key object masks, and their textual descriptions.

We conducted experiments on MLMs with 2-3 billion parameters to validate the method's ef\/fectiveness. By calculating Head Impact (HI) scores we quantify an attention head's focus on key objects, indicating its signif\/icance in image understanding. Our f\/ine-tuning experiments demonstrate that adapting layers with the highest HI scores leads to the most signif\/icant shifts in metrics compared to pre-trained, randomly selected, or lowest-HI-score layers. This indicates that f\/ine-tuning a small percentage (around 0.01\%) of parameters in these crucial layers can substantially inf\/luence image understanding capabilities.

\end{abstract}
\section{Introduction}

The development of the Transformer neural architecture \cite{vaswani2017attention} signif\/icantly accelerated progress in the f\/ield of Natural Language Processing and later, with the emergence of the Vision Transformer architecture \cite{dosovitskiy2021imageworth16x16words}, in computer vision. Modern large language models (LLMs), such as GPT-4 \cite{openai2024gpt4technicalreport}, Gemini \cite{geminiteam2024geminifamilyhighlycapable}, and LLaMA~3 \cite{grattafiori2024llama3herdmodels}, are advancing in the direction of multimodality – the simultaneous analysis of dif\/ferent data formats, such as texts and images. These multimodal language models (MLMs) have found applications in document processing tasks \cite{nacson2024docvlmmakevlmefficient}, Visual Question Answering \cite{ishman2024fromimagetolanguage}, and decision-making systems, such as autonomous driving \cite{xie2025vlmsreadyautonomousdriving}.

To adapt models to downstream tasks, f\/ine-tuning is often applied \cite{wang2025peftsurvey}. The same applies to multimodal language models, where f\/ine-tuning is also used for merging modalities \cite{caffagni2024revolution}. However, full f\/ine-tuning of a model is computationally expensive and often inaccessible to researchers. To address this problem, Parameter-Ef\/f\/icient Fine-Tuning (PEFT) methods are being developed. The essence of these methods lies in training only a small portion of the original or added model weights \cite{zhou2024empirical}, which, nevertheless, allows achieving high model quality as a result.

For ef\/fective f\/ine-tuning, it is crucial to understand which model components should undergo training to maintain a balance between the parameter-ef\/f\/iciency and the quality of the trained model. However, MLMs, like other deep learning models, are dif\/f\/icult to interpret, i.e., it is challenging to determine the inf\/luence of each model component on the decision-making process. LLMs are black boxes consisting of billions of parameters that are hard to explain. In multimodal models, the problem becomes even more complex, as new modules included, such as the Vision Encoder.

This paper proposes a method for interpreting multimodal language models, based on analyzing the model's attention scores relative to image tokens. The key idea involves identifying the attention heads of the language model that focus on key objects in the image. By key objects, we mean objects that are clearly visible in the image and contribute to its overall description. We utilize the information obtained from the interpretation method to select the most suitable model components for Parameter-Ef\/f\/icient Fine-Tuning of multimodal models.

The contribution of the work is as follows:

\begin{itemize}

    \item we propose a method for identifying the model's attention heads associated with the image key objects;

    \item we apply this method for Parameter-Ef\/f\/icient Fine-Tuning of multimodal language model layers for the Image Captioning task;

    \item we prepare a new dataset containing images, a set of key objects in these images, and textual descriptions of these objects;

    \item we conduct experiments on f\/ine-tuning MLMs with 2-3 billion parameters to validate the ef\/fectiveness of the proposed method.

\end{itemize}

\section{Related work}

\subsubsection{Multimodal Language Models.}

Modern Multimodal Language Models are based on the Transformer architecture \cite{vaswani2017attention}. MLMs consist of a visual encoder based on Vision Transformer \cite{dosovitskiy2021imageworth16x16words}, a text decoder based on Transformer, and a modality connection module \cite{yin2024mllmsurvey}. As the visual encoder, they use the ViT-part of vision-language encoder models such as CLIP \cite{radform2021learningtransferable} and SigLIP \cite{zhai2023sigmoidloss}. For the text decoder, they employ Large Language Models like LLaMA \cite{grattafiori2024llama3herdmodels}, Qwen \cite{yang2025qwen3}, and Gemma \cite{team2025gemma}.

Research on Multimodal Language Models has rapidly emerged with the open-source release of models such as LLaVA \cite{liu2023visualinstructiontuning} and BLIP2 \cite{li2023blip2}. These models employ dif\/ferent approaches to visual data representation. The LLaVA model uses an approach where images are encoded as a sequence of visual tokens, passed through a visual-language projection module, and inserted into the language model's prompt. The BLIP2 model proposed an approach with a separately trained Q-Former module, which was trained based on interactions of queries between image and text features using Self-Attention and Cross-Attention mechanisms.

In this work, we explore the LLaVA-like approach of inserting visual tokens into the language model's prompt, as it is more prevalent in modern Multimodal Language Models such as Qwen2-VL \cite{wang2024qwen2vl}, Gemma 3 \cite{team2025gemma}, and LLaVA-Next \cite{li2024llavanextstrong}. We analyze image and text tokens connections using PaliGemma 2, Qwen2-VL, and SmolVLM models.

\subsubsection{Parameter-Ef\/f\/icient Fine-Tuning for MLMs.}

Parameter-Ef\/f\/icient Fine-Tuning methods for Multimodal Language Models are used to adjust the connections between visual and linguistic modalities and to improve image understanding quality during text response generation.

In Multimodal Language Models, f\/ine-tuning can be applied to visual encoder, text decoder, and connection module. The work \cite{zhou2024empirical} presents an empirical study of using PEFT methods and MLM modules to achieve the highest performance after f\/ine-tuning.

PEFT methods are also employed to solve specif\/ic tasks. For instance, \cite{fengdi2025mhpeft} proposes a Prompt-tuning based approach MH-PEFT to address the hallucination problem in Multimodal Language Models. By f\/ine-tuning additional tokens in the language model, the authors provide additional image information, thereby reducing the likelihood of model hallucinations. Another study \cite{liu2024pefomed} suggests applying PEFT methods for f\/ine-tuning models to medical image recognition task. The authors demonstrate that f\/ine-tuning a small number of model parameters yields comparable or superior performance to specialized full-tuned models.

The core idea of our approach involves identifying the MLM parameters that have a crucial role in image understanding and subsequently tuning them. This allows for a signif\/icant reduction in the number of f\/ine-tuned parameters while simultaneously improving target task performance. Our method is not limited to specif\/ic tasks or domains and can be applied across various areas.

\subsubsection{Interpretability of Multimodal Language Models.}

The problem of interpretability in Large Language Models remains relevant \cite{brown2024enhancingtrustllmsalgorithms}. Meanwhile, an increasing number of studies focus specif\/ically on interpreting multimodal language models, investigating the connections between visual and textual tokens.

In \cite{ben2024lvlminterpret}, the authors compute attention scores between image and text tokens and visualize heatmaps of attention distribution over an image for each textual token, as well as attention distribution for textual tokens relative to selected image patches. This mechanism allows tracking specif\/ic image fragments that contribute most signif\/icantly to the answer generation process at each step.

The study \cite{neo2024interpretingvisualinformationprocessing} explores the relationship between visual and text tokens by ablating attention connections for dif\/ferent combinations of visual tokens and measuring the degree of response degradation after attentions on dif\/ferent layers are ablated. The authors propose an approach to evaluate which MLM layers have a greater impact on image analysis.

The proposed method dif\/fers in its use of attention scores to identify attention heads that focus on key objects in the image rather than the entire image. We utilize the obtained information on the inf\/luence of attention heads to investigate the possibility of Parameter-Ef\/f\/icient Fine-Tuning. In our work, we examine only the decoding language model part and do not consider attention scores of the visual encoder, as it is not involved in the answer generation process after producing visual tokens.

\section{Data}

\begin{figure}
\centering
\includegraphics[width=\linewidth]{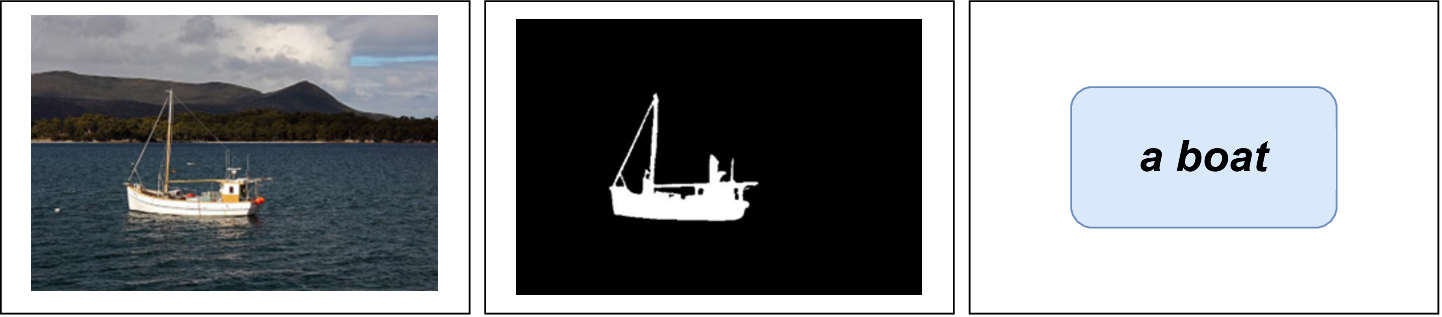}
\caption{An example of a dataset element. \textit{Left}: original image. \textit{Middle}: image key object mask. \textit{Right}: text caption of the image key object}
\label{fig_dataset_example}
\end{figure}

For investigating interpretation and conducting f\/ine-tuning experiments, we compiled a dataset where each sample contained the following elements (see Fig. \ref{fig_dataset_example}):

\begin{itemize}
    \item an image;
    \item a list of the image key objects --- set of masks highlighting each object;
    \item a list of short textual descriptions for each image key object.
\end{itemize}

We used a portion of the MS COCO 2017 validation set \cite{lin2014coco} as source images for the dataset. Each sample in this dataset contains an image and multiple general descriptions of the entire image. The provided descriptions cannot directly identify specif\/ic objects in the image. The image sizes range from (200,~145) to (640,~640). Most of the images have size (640,~480).

To obtain bounding boxes and textual descriptions of key objects, we use the Florence-2\footnote{https://huggingface.co/microsoft/Florence-2-large-ft} model \cite{xiao2024florence2}. The model was trained on the Caption to Phrase Grounding task, which allows us to identify specif\/ic object labels from a general image caption and mark them with bounding boxes. 

To obtain segmentation masks of images key objects, we use the SAM2\footnote{https://github.com/facebookresearch/sam2} model \cite{ravi2024sam2segmentimages}. The bounding boxes obtained from the Florence-2 model were fed as input to the SAM2 model to generate object masks within the provided boundaries. Key objects for which the model returned masks with conf\/idence scores below 0.85 were removed from the dataset.

The resulting dataset contains 3,000 samples. To calculate attention scores and f\/ine-tune the models, the dataset was split into two parts in a 1:2 ratio. The f\/ine-tuning dataset was then split into training (1200 samples), validation (300 samples), and test (500 samples) sets.

The dataset for calculating attention and experimenting with f\/ine-tuning is presented in the repository \footnote{https://github.com/alekosus/optimizing-mlms-icai2025}.

\section{Method}

The method is based on computing attention scores between the language response tokens and visual tokens containing information about key objects.

\begin{figure}
\centering
\includegraphics[width=\linewidth]{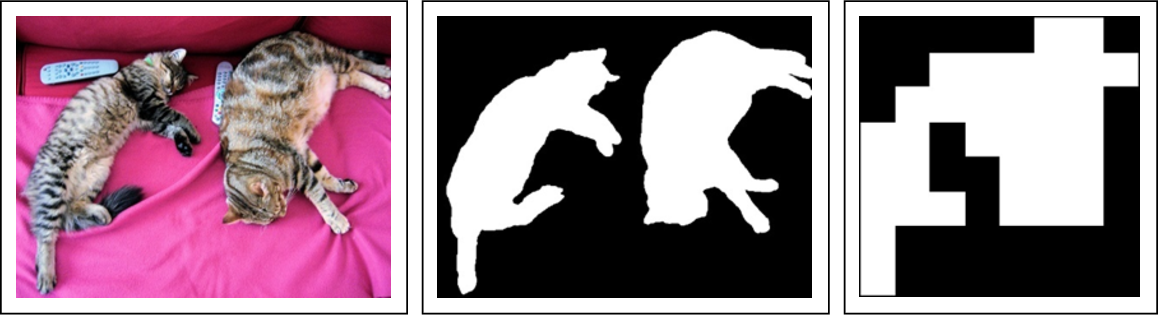}
\caption{Example of selecting key objects in an image. \textit{Left}: original image. \textit{Middle}: image key object mask. \textit{Right}: representation of the mask as a sequence of visual tokens with f\/ixed size image patches}
\label{fig_method_image_processing}
\end{figure}

An example of key object extraction from an image is shown in Figure \ref{fig_method_image_processing}. Key objects in the image are represented by a mask (see Fig. \ref{fig_method_image_processing}, Middle) – a matrix with the same shape as the original image, containing values of 1 and 0 indicating whether a pixel belongs to the key object or not. The binary mask of the key object contains in dataset and undergoes the same transformations as the input image to generate visual tokens, thereby establishing correspondence between pixels in the object mask and their corresponding visual tokens. The image mask in the form of visual tokens is represented as a matrix with shape (\texttt{visual tokens rows, visual tokens columns}) (see Fig. \ref{fig_method_image_processing}, Right).

To compute attention scores, we process the data through the model on the Image Captioning task. Specif\/ic prompts used in the experiments are provided in the Appendix. In the User prompt, we insert the image along with instructions to describe it, while the Assistant prompt contains descriptions of the image key objects taken from the dataset.

\begin{figure}
\centering
\includegraphics[width=0.99\textwidth]{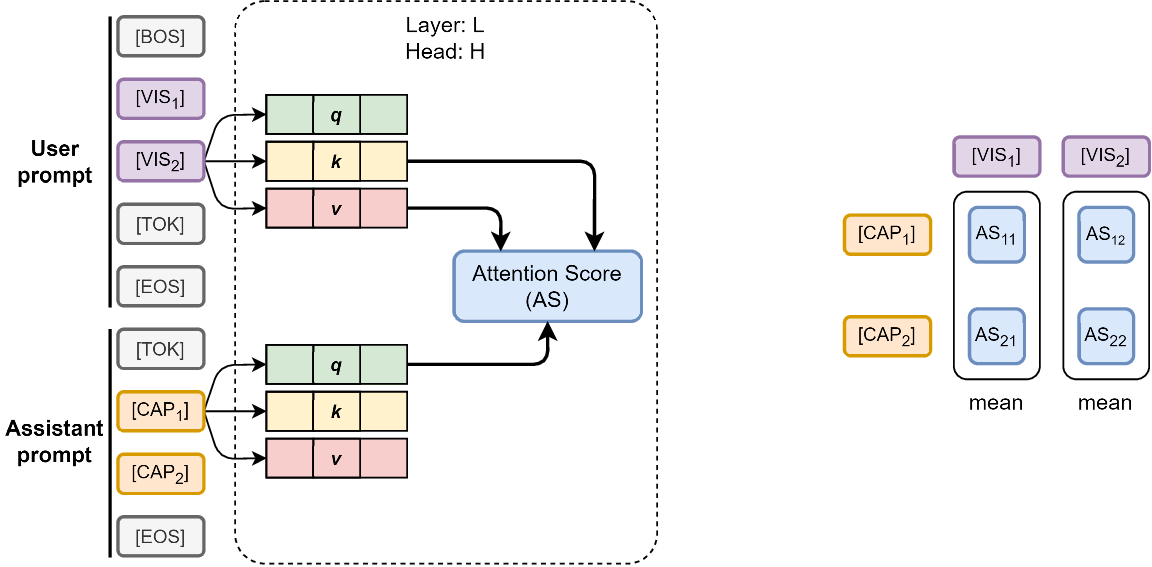}
\caption{The architecture of calculating attention scores. \textit{Left}: Representation of the attention mechanism within an LLM. To calculate attention scores, queries are taken from the object caption tokens ([CAP]) in the Assistant prompt and keys and values from the visual tokens ([VIS]) in the User prompt. \textit{Right}: The process of computing average scores for an image token. The calculated attention scores (AS) are averaged for each caption token from the Assistant prompt.} 
\label{fig_method_attention}
\end{figure}

The architecture for computing attention scores is shown in Figure \ref{fig_method_attention} and refers to the Self-Attention mechanism introduced in \cite{vaswani2017attention}. The model prompt contains sequences of image tokens, which encode image information, and response language tokens, which contain descriptions of key objects. We use the response tokens as queries and the image tokens as keys and values, then compute Self-Attention between them. Since the response text may consist of multiple tokens, the resulting attention scores for all response tokens are averaged. This produces an attention score for each image token.

Attention scores are computed for each attention head in each layer of the language model for every visual token. Thus, the score tensor has shape (\texttt{number of layers, number of heads, visual tokens rows, visual tokens columns}). The attention scores were converted to binary values to establish correspondence with object masks. For each specif\/ic head and layer, all scores above the mean across all image tokens were set to 1, while those below the mean were set to 0.

The binarization of attention score matrices and masks enables us to overlay them and compute their correspondence. We utilize Intersection over Union (IoU, also known as Jaccard Index) (\ref{eq_method_iou}) between attention scores and corresponding token mask values to measure a specif\/ic attention head's focus on image key objects (here in the formula (\ref{eq_method_iou}), the symbol $\odot$ denotes element-wise matrix multiplication, $[comparison]$ denotes Iverson bracket --- boolean to numeric conversion: \textit{true} represents 1, \textit{false} represents 0). IoU yields values in the range [0, 1], where 1 indicates complete overlap between attention and mask, while 0 signif\/ies no overlap.

\begin{equation}
\label{eq_method_iou}
IoU(l, h)=\frac{\sum_{x,y} Attentions(l,h) \odot Mask}{\sum_{x,y} [(Attentions(l,h) + Mask) > 0]}
\end{equation}

To compute the overall Head Impact (HI) score assessing a specif\/ic head's inf\/luence relative to image key objects, IoU scores are averaged across the entire dataset $D$ (\ref{eq_method_hi}).

\begin{equation}
\label{eq_method_hi}
HI(l,h) = \frac{1}{|D|}\sum_{d \in D} IoU_{d}(l,h)
\end{equation}

A high Head Impact value indicates that the attention head focuses more on key objects during response generation (which contains descriptions of image key objects). This can be interpreted as heads with high Head Impact having greater signif\/icance in image understanding. 

We use information about the inf\/luence of attention heads to select parameters for Parameter-Ef\/f\/icient Fine-Tuning. We verify that f\/ine-tuning attention heads with high Head Impact scores leads to a greater change in model performance than f\/ine-tuning heads with low Head Impact score or randomly selected attention heads.

\section{Experiments}

\subsection{Models}

For our experiments, we selected three multimodal generative language models with approximately 2-3 billion parameters: PaliGemma2-3B-Mix\footnote{https://huggingface.co/google/paligemma2-3b-mix-448}, Qwen2-VL-2B-Instruct\footnote{https://huggingface.co/Qwen/Qwen2-VL-2B-Instruct}, and SmolVLM-Instruct\footnote{https://huggingface.co/HuggingFaceTB/SmolVLM-Instruct} (2 billion parameters). These models were chosen due to their dif\/ferences in visual token sequence representations.

PaliGemma2 model \cite{steiner2024paligemma2familyversatile} uses the traditional approach of representing an image using a projection of tokens from an image encoder. The image is scaled to a f\/ixed resolution (448,~448) pixels and encoded with inner patch size (14,~14). In this case, one image is represented by a f\/ixed length sequence of 1,024 visual tokens.

PaliGemma2 was trained on specif\/ic prompt templates, so the prompts used in experiments with this model are dif\/ferent from others.

The authors of Qwen2-VL model \cite{wang2024qwen2vl}, in contrast to PaliGemma2, encode the image into a sequence of visual tokens of variable length depending on the resolution. Images are scaled to preserve the approximate original resolution, encoded using a modif\/ied version of ViT with inner patches (14,~14) and the each 4 tokens from encoder merges into 1 visual token.

The SmolVLM model is based on the Idef\/ics3 model architecture \cite{laurençon2024buildingbetterunderstandingvisionlanguage}. The input image is scaled with aspect ratio preservation and then cropped into multiple sub-images with f\/ixed resolution (384,~384) pixels. In this way, images with larger resolution are represented by more sub-images and thus more visual tokens. To retain the global context of the image, a scaled original image is also included in the prompt. Each sub-image is then encoded with inner patches (14,~14) and every 9 tokens from encoder merges into 1 visual token. Each sub-image is encoded to 81 visual tokens.

When computing attention scores for image tokens in the SmolVLM model, we considered attention to both objects in cropped fragments and the complete image.

A single image token corresponds to a 14×14 patch for PaliGemma2, a 28×28 patch for Qwen2.5-VL, and a 42×42 patch for SmolVLM.

\subsection{Calculating Head Impact}

\begin{figure}
\centering
\includegraphics[width=\linewidth]{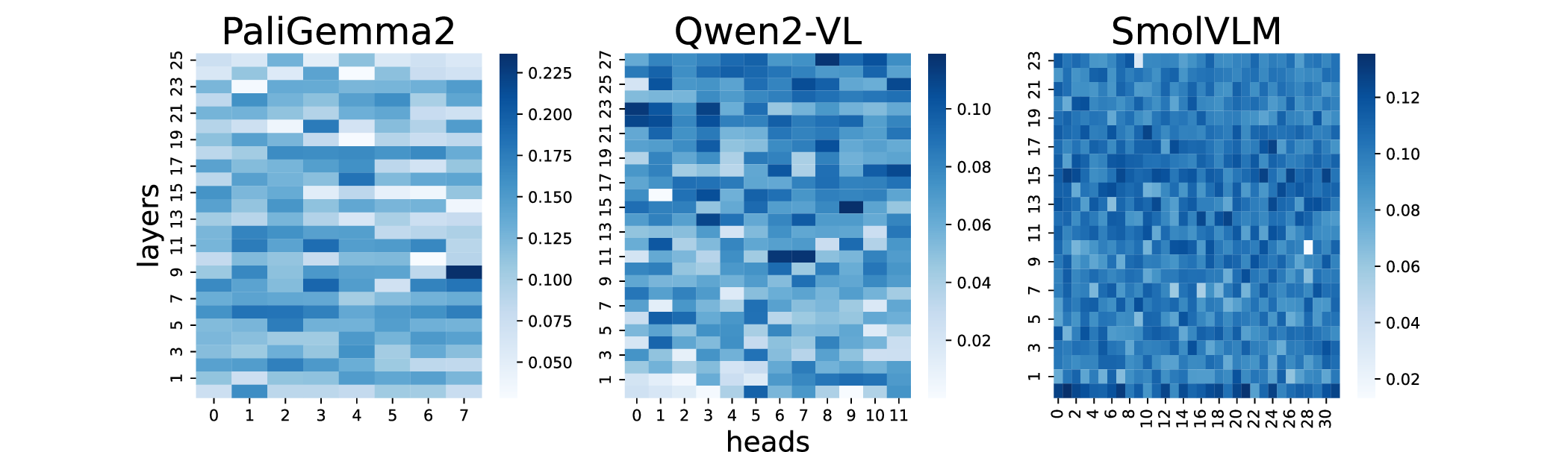}
\caption{Head Impact scores for the models.}
\label{fig_experiments_headimpacts}
\end{figure}

We compute Head Impact scores for the examined MLMs. The model conf\/igurations are presented in Table~\ref{tab_experiments_model_stats}. The heatmap of HI scores is shown in Figure~\ref{fig_experiments_headimpacts}.

\begin{table}
\centering
\caption{Conf\/igurations of MLLMs.}
\label{tab_experiments_model_stats}
\begin{tabular}{|l|r|r|r|}
\hline
\multicolumn{1}{|l|}{Model} & \multicolumn{1}{l|}{\# parameters} & \multicolumn{1}{l|}{Layers} & \multicolumn{1}{l|}{Attention heads} \\
\hline
PaliGemma2 & 3.03B & 26 & 8\\
Qwen2-VL & 2.21B & 28 & 12\\
SmolVLM & 2.25B & 24 & 32\\
\hline
\end{tabular}
\end{table}

\begin{figure}
\centering
\includegraphics[width=0.99\linewidth]{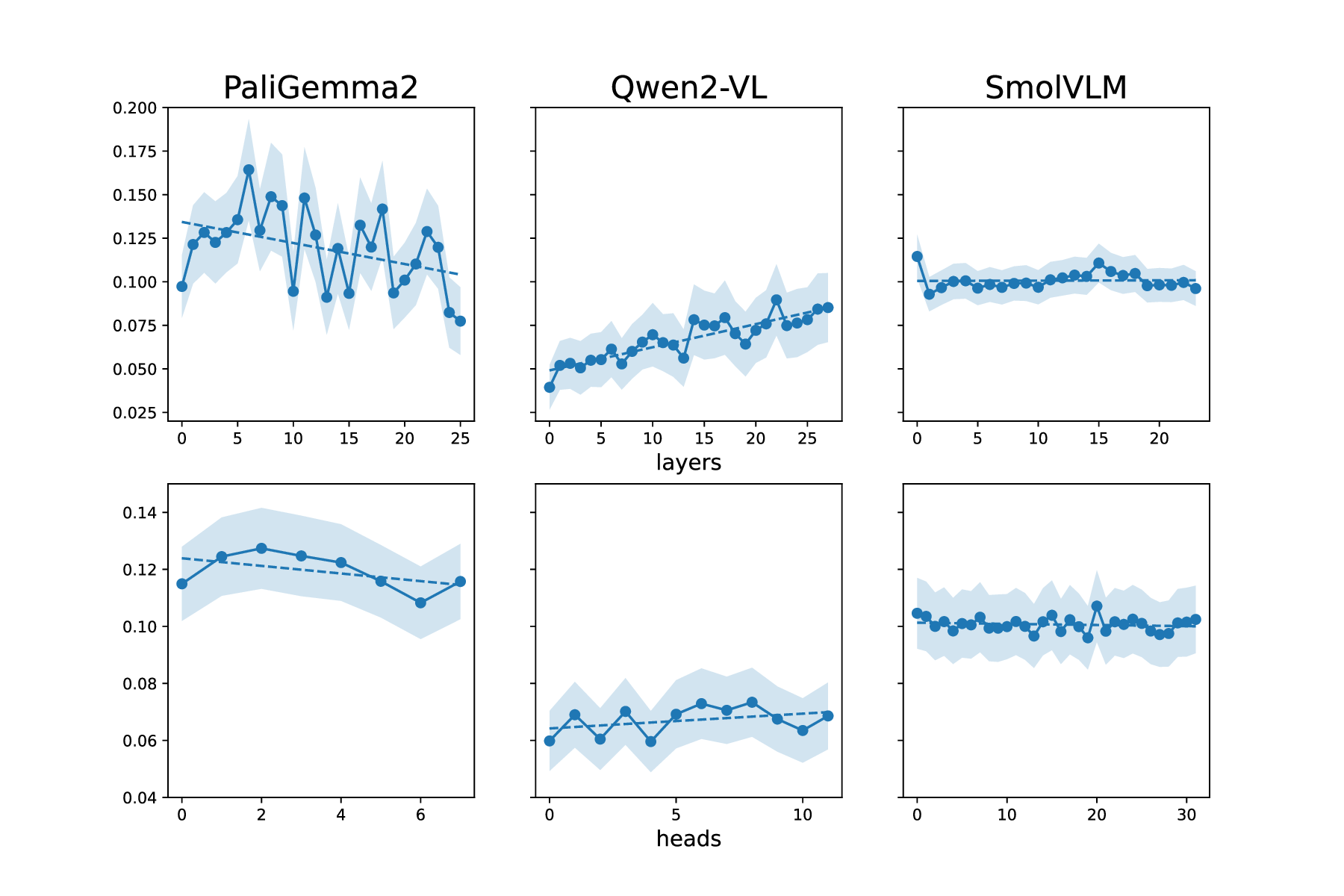}
\caption{Average Head Impact scores. Solid line with dots --- scores for the layer/head, dashed line --- regression line, f\/illed area --- standard error. \textit{Top}: layer-wise averaged scores. \textit{Bottom}: head-wise averaged scores.}
\label{fig_experiments_himeans}
\end{figure}

We observed that Head Impact scores vary across layers, while remaining approximately consistent among attention heads within the same layer (see Figure~\ref{fig_experiments_himeans}). To validate these hypotheses, we applied the Kruskal-Wallis statistical test. The test results are presented in~Table \ref{tab_experiments_stats} (p-values < 0.001 are highlighted). The statistical test conf\/irmed the hypothesis of statistically signif\/icant dif\/ferences in HI scores across layers and rejected the hypothesis of statistically signif\/icant dif\/ferences in HI scores across heads.

\begin{table}
\centering
\caption{Kruskal--Wallis test on Head Impact scores.}
\label{tab_experiments_stats}
\begin{tabular}{|l|r|r|r|r|r|r|}
\hline
& \multicolumn{2}{c|}{PaliGemma2} & \multicolumn{2}{c|}{Qwen2-VL} & \multicolumn{2}{c|}{SmolVLM}\\
& \multicolumn{1}{c|}{Layers} & \multicolumn{1}{c|}{Heads} & \multicolumn{1}{c|}{Layers} & \multicolumn{1}{c|}{Heads} & \multicolumn{1}{c|}{Layers} & \multicolumn{1}{c|}{Heads}\\
\hline
statistic & 92.93 & 12.94 & 103.56 & 31.86 & 75.28 & 6.81\\
p-value & $\mathbf{3.62 \cdot 10^{-09}}$ & 0.30 & $\mathbf{3.40 \cdot 10^{-12}}$ & 0.42 & $\mathbf{6.15 \cdot 10^{-07}}$ & 0.449\\
\hline
\end{tabular}
\end{table}

\subsection{Experimental setup}

We compare the following model setups:

\begin{itemize}
    \item original model --- the model without f\/ine-tuning of attention heads;
    \item top-4 --- the model where four layers with the highest Head Impact scores were f\/ine-tuned;
    \item bottom-4 --- the model where four layers with the lowest Head Impact scores were f\/ine-tuned;
    \item random-4 --- the model where four randomly selected layers, not included in top-4 or bottom-4, were f\/ine-tuned;
    \item full --- the model where all layers were f\/ine-tuned.
\end{itemize}

We perform f\/ine-tuning using LoRA modules \cite{hu2021loralowrankadaptationlarge} attached to the query, key, and value weight matrices of the language model's attention blocks. LoRA hyperparameters: $r=8$, $\alpha=16$, without dropout and biases. Table \ref{tab_experiments_layers} shows the indices of the layers that were f\/ine-tuned for each model setup.

\begin{table}
\centering
\caption{Indices of the trainable layers for each experiment.}
\label{tab_experiments_layers}
\begin{tabular}{|p{3cm}|p{3cm}|}
\hline
\multicolumn{1}{|c|}{setup} & \multicolumn{1}{|c|}{layers}\\
\hline
\multicolumn{2}{|l|}{PaliGemma2}\\
\hline
top-4 & 6, 8, 9, 11\\
bottom-4 & 13, 15, 24, 25\\
random-4 & 5, 7, 10, 16\\
\hline
\multicolumn{2}{|l|}{Qwen2-VL}\\
\hline
top-4 & 17, 22, 26, 27\\
bottom-4 & 0, 1, 3, 7\\
random-4 & 2, 13, 14, 24\\
\hline
\multicolumn{2}{|l|}{SmolVLM}\\
\hline
top-4 & 0, 15, 16, 18\\
bottom-4 & 1, 2, 5, 23\\
random-4 & 3, 7, 8, 17\\
\hline
\end{tabular}
\end{table}

\subsection{Task description}

We perform f\/ine-tuning on the Image Captioning task. The User prompt contains an image along with instructions to describe it. The Assistant prompt includes a description of the image key object from the dataset.

To evaluate the Image Captioning performance, we analyze the probability distribution of output tokens in prepared test data containing both User and Assistant prompts. The idea behind answer substitution lies in eliminating response redundancy that might occur if the model generated answers autonomously. We employ Perplexity score over image key object description tokens to analyze the model's expectation level of the generated sequence (\ref{eq_peplexity}). Perplexity measures how well the language model predicts the token distribution in responses; calculated as the exponential of the average negative log-probability of the response tokens~$T$.

\begin{equation}
    \label{eq_peplexity}
    Perplexity = \exp{\left(\frac{1}{N} \sum_{N} \frac{1}{T} \sum_{t \in T}(-\log p_t)\right)}
\end{equation}

We also evaluate the f\/ine-tuned model on the closed-ended Visual Question Answering task. The prompt for this task includes four answer options --- one valid description of the key object in the image and three false descriptions randomly selected from the dataset. To determine the model's answer, the probabilities of the answer label tokens are calculated: A, B, C, or D, and the answer with the highest probability of generating the corresponding token is selected.

To assess model performance on the closed-ended Visual Question Answering task, we use the classic Accuracy metric --- total proportion of correctly selected answer options; calculated as the ratio of correctly indicated answers to all answers.

We evaluate model performance on the Visual Question Answering task without explicitly adding this task's data to the training set, which, according to the results, leads to performance degradation on the task. This demonstrates the sensitivity of the model's trained layers and supports the idea that selecting specif\/ic layers can signif\/icantly impact learning ability.

The task prompts for each model are provided in the Appendix.

\subsection{Results}

We calculated metrics using 500 test data samples. Table \ref{tab_experiments_ftmetrics} presents the metric values. For the Perplexity metric, lower values indicate better performance, while for the Accuracy metric, higher values indicate better performance. A larger deviation from the original setup indicates a greater impact of layer selection during f\/ine-tuning.

\begin{table}
\centering
\caption{Fine-tuning metric values with standard errors. The largest absolute dif\/ferences in metrics compared to the original model are \textbf{highlighted}.}
\label{tab_experiments_ftmetrics}
\begin{tabular}{|l|r|r|r|r|r|}
\hline
& original & top-4 & bottom-4 & random-4 & full\\
\hline
\multicolumn{6}{|l|}{PaliGemma2}\\
\hline
Accuracy $\uparrow$ & 0.94 & \textbf{0.26} & 0.94 & 0.94 & 0.90\\
AnsProb $\uparrow$ & 0.94 $ \pm $ 0.01 & \textbf{0.26 $ \pm $ 0.01} & 0.94 $ \pm $ 0.01 & 0.94 $ \pm $ 0.01 & 0.91 $ \pm $ 0.01\\
Perplexity $\downarrow$ & 9.85 $ \pm $ 0.04 & \textbf{8.88 $ \pm $ 0.08} & 9.85 $ \pm $ 0.04 & 9.85 $ \pm $ 0.04 & 9.46 $ \pm $ 0.04\\
\hline
\multicolumn{6}{|l|}{Qwen2-VL}\\
\hline
Accuracy $\uparrow$ & 0.95 & \textbf{0.92} & 0.95 & 0.95 & 0.93\\
AnsProb $\uparrow$ & 0.93 $ \pm $ 0.01 & \textbf{0.86 $ \pm $ 0.01} & 0.93 $ \pm $ 0.01 & 0.93 $ \pm $ 0.01 & 0.88 $ \pm $ 0.01\\
Perplexity $\downarrow$ & 13.74 $ \pm $ 0.13 & \textbf{12.89 $ \pm $ 0.12} & 13.74 $ \pm $ 0.13 & 13.74 $ \pm $ 0.13 & 13.30 $ \pm $ 0.13\\
\hline
\multicolumn{6}{|l|}{SmolVLM}\\
\hline
Accuracy $\uparrow$ & 0.91 & \textbf{0.81} & 0.85 & 0.91 & 0.89\\
AnsProb $\uparrow$ & 0.84 $ \pm $ 0.01 & \textbf{0.71 $ \pm $ 0.01} & 0.76 $ \pm $ 0.01 & 0.84 $ \pm $ 0.01 & 0.87 $ \pm $ 0.01\\
Perplexity $\downarrow$ & 17.20 $ \pm $ 0.10 & \textbf{15.39 $ \pm $ 0.10} & 16.04 $ \pm $ 0.08 & 17.20 $ \pm $ 0.10 & 16.77 $ \pm $ 0.10\\
\hline
\end{tabular}
\end{table}

For all models, f\/ine-tuning the four layers with the highest Head Impact scores (top-4) shows the most signif\/icant metric dif\/ferences compared to pre-trained models. For models where other layers were f\/ine-tuned (bottom-4, random-4), the metrics after f\/ine-tuning demonstrate much smaller dif\/ferences.

\section{Discussion}

Experiments reveal the presence of attention heads that focus on specif\/ic key objects in images. These heads can be identif\/ied through analysis of attention scores between response language tokens and visual tokens in the MLM prompt using the Head Impact scores.

It should be noted that attention heads within the same layer demonstrate similar attention patterns regarding key objects. However, heads across dif\/ferent layers do not maintain this consistent behavior. Therefore, the concept of attention heads with high understanding of image key objects can be generalized to the layer level. The Kruskal-Wallis test demonstrates statistical signif\/icance in dif\/ferences of the mean Head Impact scores across layers. Thus, it can be assumed that individual Transformer blocks are more trained to pay attention to image key objects.

At the same time, no consistent pattern is observed in the positions of layers with higher HI scores (see Fig. \ref{fig_experiments_himeans}). For PaliGemma2 model, the average Head Impact scores across layers decrease, while for Qwen2-VL model they, on the contrary, increase, which can be observed from the regression line.

Fine-tuning layers with the highest Head Impact scores (top-4) has the most signif\/icant ef\/fect on metric dif\/ferences, which may be attributed to their crucial role in processing visual object information. High HI scores indicate that the corresponding attention heads focus on semantically signif\/icant image elements, enabling more informed generation of textual descriptions. Strengthening these weights through f\/ine-tuning increases the model's sensitivity to relevant visual tokens, leading to more substantial shifts in token distributions within MLMs.

However, as demonstrated by f\/ine-tuning experiments on the Visual Question Answering task, this weight sensitivity also implies forgetting of pre-trained knowledge when downstream task data is excluded during f\/ine-tuning. In the case of PaliGemma2, which encodes images into f\/ixed-length token sequences, the Accuracy drop before and after f\/ine-tuning the top four layers reaches a substantial 0.676.

In contrast, f\/ine-tuning layers with the lowest Head Impact scores or random layers does not demonstrate signif\/icant dif\/ferences in metrics. We can hypothesize that the attention heads in these layers do not focus on the image key objects but rather perform other functions. Consequently, f\/ine-tuning these layers does not modify the model's internal attention structure to improve performance on multimodal tasks.

The metric dif\/ferences may stem from how visual information is represented in the MLM prompt. For instance, the PaliGemma2 model, where the visual token sequence always has a f\/ixed length and is formed through linear projection without merging, exhibits the most substantial Accuracy shift in the Visual Question Answering task. In the case of Qwen2-VL and SmolVLM models, token aggregation and variable count of visual tokens depending on input image resolution mitigate the ef\/fect of individual attention heads, reducing metric dif\/ferences during f\/ine-tuning.

Table \ref{tab_discussion_modelparams} presents statistics on the number of trainable parameters for each model (weight matrices for queries, keys, and values across four layers). Thus, training only 0.01\% of MLM parameters can inf\/luence image understanding capabilities.

\begin{table}
\centering
\caption{Parameter statistics in Parameter-Ef\/f\/icient Fine-Tuning Experiments.}
\label{tab_discussion_modelparams}
\begin{tabular}{|l|r|r|r|}
\hline
model & total, \#  & trainable, \# & trainable, \%\\
\hline
PaliGemma2 & 3,033,479,408 & 352,256 & 0.0116\\
Qwen2-VL & 2,209,198,592 & 212,992 & 0.0096\\
SmolVLM & 2,246,666,096 & 393,216 & 0.0175\\
\hline
\end{tabular}
\end{table}

\section{Conclusion}

The paper proposes a method for interpreting attention heads of Multimodal Language Models and investigates its application for Parameter-Ef\/f\/icient Fine-tuning. The method enables identif\/ication of attention heads in Transformer-based models that focus on key objects in images. We compute an overall impact score for image understanding by specif\/ic attention heads – Head Impact score. The Kruskal-Wallis statistical test demonstrates the existence of layers in MLMs that have greater inf\/luence on image understanding.

We conduct experiments with PaliGemma2, Qwen2-VL and SmolVLM models to compute Head Impact scores and perform parameter-ef\/f\/icient f\/ine-tuning on the Image Captioning task. The experiments show that f\/ine-tuning adapters for specif\/ic model layers with the best Head Impact scores yields the most signif\/icant metric shifts compared to the pre-trained model.

\begin{credits}
\subsubsection{Limitations.} 
In our work, experiments are limited to models with similar architectures (Vision Transformer encoder and Transformer decoder) and the method of representing images as visual tokens embedded in the language model prompt. We are also constrained by the model sizes of 2-3 billion parameters due to computational limitations.

Although MLMs are generative models, we did not conduct experiments with text sequence generation. Instead, we inserted answer templates into the model's assistant prompt. This approach was chosen to eliminate ambiguity in generated responses and enable unambiguous computation of attention scores and metrics. The templates were constructed based on actual model responses to avoid output token distribution shifts and consequent performance degradation. We plan to evaluate the f\/ine-tuning method on open-ended generative tasks in future work.

\subsubsection{\discintname}
The authors have no competing interests to declare that are relevant to the content of this article.
\end{credits}

\section*{Appendix}

Template prompts for MLLMs used in the experiments.

Template prompts for calculating Head Impact scores:

\begin{itemize}
    \item For Qwen2-VL and SmolVLM models:

    \medskip
    \begin{minipage}{\linewidth}
    \centering
    \begin{tabular}{r p{5cm}}
    \textbf{User:} & <Image>\\
    & What is this photo?\\
    \textbf{Assistant:} & This is a photo of \{description\}.\\
    \end{tabular}
    \end{minipage}
    \medskip

    \item For PaliGemma2 model ("en" means "in English"):

    \medskip
    \begin{minipage}{\linewidth}
    \centering
    \begin{tabular}{r p{5cm}}
    \textbf{User:} & <Image>\\
    & describe en\\
    \textbf{Assistant:} & \{description\}\\
    \end{tabular}
    \end{minipage}
    
\end{itemize}

Template prompt for the closed-ended question answering task:

\begin{itemize}
    \item For Qwen2-VL and SmolVLM models:

    \medskip
    \begin{minipage}{\linewidth}
    \centering
    \begin{tabular}{r p{5cm}}
    \textbf{User:} & <Image>\\
    & What is shown in this photo? Choose one of the options:\\
    & A) ... B) ... C) ... D) ...\\
    \textbf{Assistant:} & \{A-D\}.\\
    \end{tabular}
    \end{minipage}
    \medskip

    \item For PaliGemma2 model ("en" means "in English"):

    \medskip
    \begin{minipage}{\linewidth}
    \centering
    \begin{tabular}{r p{5cm}}
    \textbf{User:} & <Image>\\
    & answer en\\
    & A) ... B) ... C) ... D) ...\\
    \textbf{Assistant:} & \{A-D\}.\\
    \end{tabular}
    \end{minipage}

\end{itemize}

Prompt for the open-ended question answering task is the same as the prompt for calculating Head Impact scores for all models.

\bibliographystyle{splncs04}
\bibliography{mybibliography}

\end{document}